# FACE RECOGNITION WITH SMALL AND LARGE SIZE DATABASES[1]

**Josep Roure, Marcos Faundez-Zanuy**
**Escola Universitària Politècnica de Mataró, Adscrita a la UPC**
e-mail: {roure, faundez}@eupmt.es

## ABSTRACT

This Paper will present experimental results using the ORL (40 people) and FERET (994 people) databases. The ORL database can be useful for securing applications where few users attempting to access are expected. This is the case, for instance, of a PDA or PC where the password is the face of the user. On the other hand the FERET database is useful for studying those situations where the number of authorized users is around a thousand people.

**Keywords**: Face image, Discrete Cosine Transform, verification, feature extraction.

## 1. INTRODUCTION

Face recognition is one of the most successful ways for biometric recognition. In [1] we presented a Discrete Cosine Transform (DCT) approach for face recognition. That approach outperformed the classical eigenface algorithm [2] and reduced the computational burden. In this paper, we will extend our previous results in several ways:

a) We will use a larger database: FERET [3].
b) We will study a verification application. That is: the system must check if the user is who he/her claims to be.
c) We will study the relevance of colour information.

The use of a small-size database is interesting for developing algorithms, because experimental simulations are much less intensive. Additionally, its worth is remarkable for those scenarios where few users are expected, such as accessing a Personal computer or a PDA without providing any identity (PIN-less access or matching based on 1:N comparisons). Theoretically, for a large set of users, PIN-less mode is not recommended [4], because False Match Errors will be high.

For verification applications, the database size should not present any relevance, because each matching is based on 1:1 comparison. However, in practice, the obtained results are much more statistical significant when using larger databases, especially for low error rates. In [5] the minimum size of the test data set, $N$, which guarantees statistical significance in a pattern recognition task, is derived. The goal in the abovementioned work is to estimate $N$ so that it is guaranteed, with a risk $\alpha$ of being wrong, that the error rate $P$ does not exceed that estimated from the test set, $\hat{P}$, by an amount larger than $\varepsilon(N,\alpha)$, that is,

$$\Pr\{P > \hat{P} + \varepsilon(N,\alpha)\} < \alpha \qquad (1)$$

Letting $\varepsilon(N,\alpha) = \beta P$ and supposing recognition errors as Bernoulli trials (i.i.d. errors), we can derive the following relation after some approximations:

$$N \approx \frac{-\ln \alpha}{\beta^2 P} \qquad (2)$$

For typical values of $\alpha$ and $\beta$ ($\alpha$ =0.05 and $\beta$ =0.2), the following simplified criterion is obtained:

$$N \approx \frac{100}{P} \qquad (3)$$

If the samples in the test data set are not independent (due to correlation factors that may include variations in recording conditions, in the type of sensors, etc.), then $N$ must be further increased. The reader is referred to [5] for a detailed analysis of this case, where some guidelines for computing the correlation factors are also given.

## 2. ORL DATABASE

The first used database is the ORL (Olivetti Research Laboratory) faces database [6]. This database contains a set of face images taken between April 1992 and April 1994 at ORL. The database was used in the context of a face recognition project carried out in collaboration with the Speech, Vision and Robotics Group of the Cambridge University Engineering Department.

There are ten different images of each of 40 distinct subjects. For some subjects, the images were taken at different times, varying the lighting, facial expressions (open/closed eyes, smiling / not smiling) and facial details (glasses / no glasses). All the images were taken against a dark homogeneous background with the subjects in an upright, frontal position (with tolerance for some side movement).

**Conditions of the experiments**
Our results have been obtained with the ORL database in the following situation: 40 people, faces 1 to 5 for training, and faces 6 to 10 for testing.

We obtain one model from each training image. During testing each input image is compared against all the models inside the database (40x5=200 in our case) and the model close to the input image (using Mean Square Error criterion) indicates the recognized person. In our experiments, we are making for each user, all other

---
[1] This work has been supported by FEDER and the Spanish grant MCYT TIC2003-08382-C05-02

users' samples as impostor test samples, so we finally have, that $N = 40 \times 5$(client) $+ 40 \times 39 \times 5$ (impostors) = 8000. So, with 95% confidence, our experiments guarantee statistical significance in experiments with an empirical error rate $\hat{P}$, down to 1.25%, which is certainly suitable for our experiments.

Figure 1 shows some samples of ORL database.

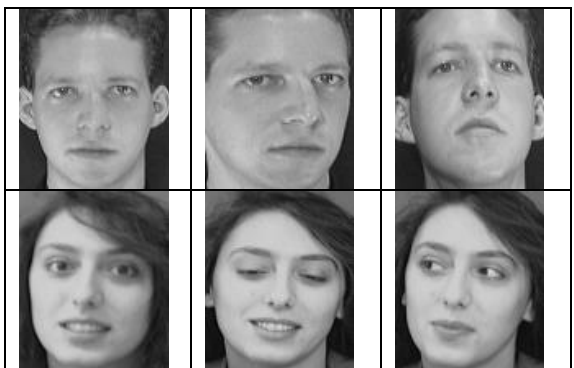

*Figure 1. Example of ORL face images*

## 3. FERET DATABASE

In our experiments, we are making for each user, all other users' samples as impostor test samples, so we finally have, that $N$=992 (client) + 993×992 (impostors) = 986048. So, with 95% confidence, our experiments guarantee statistical significance in experiments with an empirical error rate $\hat{P}$, down to 0.01%.

Figure 2 shows some samples of FERET database.

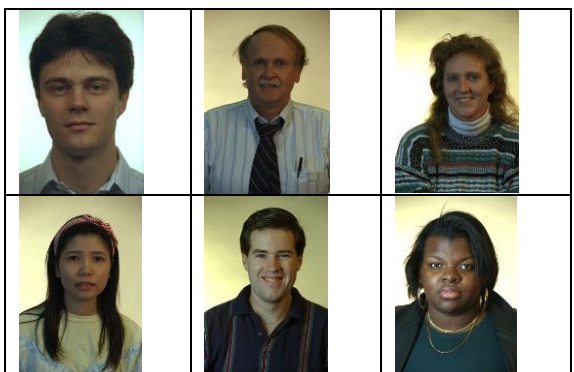

*Figure 2. Example of FERET face images.*

## 4. EXPERIMENTAL RESULTS

Biometric recognition systems can be operated in two different modes: Identification and verification.

**Identification**
Figure 3 shows the general scheme of a biometric system. These systems can be operated in two ways:
a) Identification: In this approach no identity is claimed from the person. The automatic system must determine who is trying to access.
b) Verification: In this approach the goal of the system is to determine whether the person is who he/she claims to be. This implies that the user must provide an identity and the system just accepts or rejects the users according to a successful or unsuccessful verification. Sometimes this operation mode is named authentication or detection.

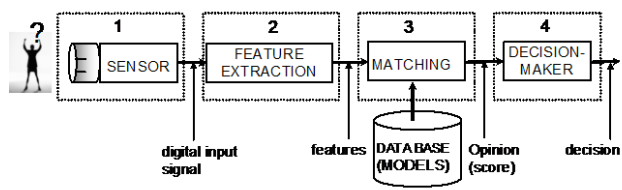

*Figure 3. General scheme of a biometric recognition system.*

For identification, if we have a population of *N* different people, and a labelled test set, we fill up a matrix *S*, where the elements are interpreted in the following way:

$$s_{ijk} = O[j]\big|_{\vec{x} \in person\#i} \quad, k=1,\cdots \#trials \quad (4)$$

Where trials is the number of different testing images per person (in our experiments, #*trials*=5 for ORL and 1 for FERET), and $s_{ijk}$ is the similarity from the *k* realization of an input signal belonging to person *i*, to the model of person *j*.

This matrix can be drawn as a three dimensional data structure (see figure 4).

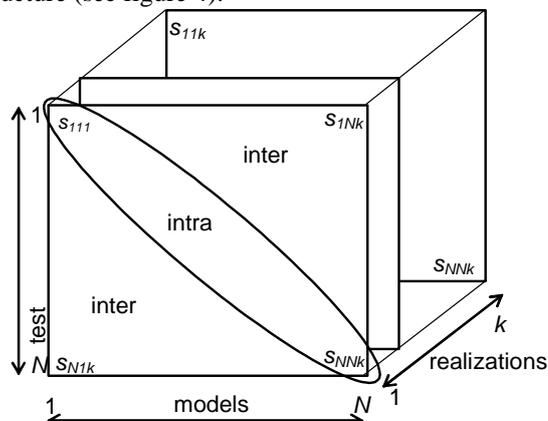

*Figure 4. Proposed data structure.*

Thus, the identification rate looks for each realization, in each raw, if the maximum similarity is inside the principal diagonal (success) or not (error), and works out the identification rate as the ratio between successes and number of trials (successes + errors):

```
for i=1:N,
    for k=1:#trials,
        if(s_iik>s_ijk)  ∀j≠i, then
            success=success+1
        else error=error+1
        end
    end
end
```

**Verification**

Verification systems can be evaluated using the False Acceptance Rate (FAR, those situations where an impostor is accepted) and the False Rejection Rate (FRR, those situations where a user is incorrectly rejected), also known in detection theory as False Alarm and Miss, respectively. There is trade-off between both errors, which has to be usually established by adjusting a decision threshold. The performance can be plotted in a ROC (Receiver Operator Characteristic) or in a DET (Detection error trade-off) plot [10]. DET curve gives uniform treatment to both types of error, and uses a logarithm scale for both axes, which spreads out the plot and better distinguishes different well performing systems and usually produces plots that are close to linear. DET plot uses a logarithmic scale that expands the extreme parts of the curve, which are the parts that give the most information about the system performance. For this reason the speech community prefers DET instead of ROC plots. Figure 5 shows an example of DET plot.

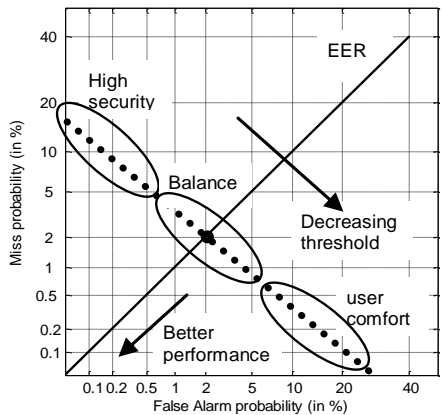

*Figure 5. Example of a DET plot for a user verification system (dotted line). The Equal Error Rate (EER) line shows the situation where False Alarm equals Miss Probability (balanced performance). Of course one of both errors rates can be more important (high security application versus those where we do not want to annoy the user with a high rejection/ miss rate). If the system curve is moved towards the origin, smaller error rates are achieved (better performance). If the decision threshold is reduced, we get higher False Acceptance/Alarm rates.*

We have used the minimum value of the Detection Cost Function (DCF) for comparison purposes. This parameter is defined as [10]:

$$DCF = C_{miss} \times P_{miss} \times P_{true} + C_{fa} \times P_{fa} \times P_{false} \quad (5)$$

Where $C_{miss}$ is the cost of a miss (rejection), $C_{fa}$ is the cost of a false alarm (acceptance), $P_{true}$ is the a priori probability of the target, and $P_{false} = 1 - P_{true}$, $C_{miss} = C_{fa} = 1$.

Using the data structure defined in figure 4, we can easily apply the DET curve analysis. We just need to split the distances into two sets: intra-distances (those inside the principal diagonal), and inter-distances (those outside the principal diagonal).

**Face Recognition**

Face recognition is probably the most natural way to perform a biometric authentication between human beings. Face recognition can rely on single still images, multiple still images, or video sequence. Although traditionally most efforts have been devoted to the former one, the latest ones are quickly emerging [7], probably due to the reduction of price in image and video acquisition devices. For instance, a sequence of images can provide a unimodal data fusion scheme [8], where the verification relies on a set of images, rather than on a single one.

Two main approaches exist for face recognition:
a) Statistical approaches consider the image as a high dimensional vector, where each pixel is mapped to a component of a vector. Due to the high dimensionality of the vectors some vector dimension reduction algorithm must be used. Typically the Karhunen Loeve transform is applied [2].
b) Geometry-feature-based methods try to identify the position and relationship between face parts, such as eyes, nose, mouth, etc., and the extracted parameters are measures of textures, shapes, sizes, etc. of these regions.

In this paper, we will focus on the first approach.

The matching techniques can be split into three categories: template matching methods, stochastic methods, and neural networks:
1. <u>Template matching methods:</u> The input and model faces are expressed as feature vectors and compared using a distance measure between them.
2. <u>Statistical methods:</u> The features extracted from the training faces are used to work out a statistical model. During testing, the similarity of input and reference is established. The most popular methods belonging to this category are Hidden Markov Models (HMM), Gaussian Mixture Models (GMM), etc.
3. <u>Neural Networks:</u> For instance, a Multi-Layer Perceptron can perform as a classifier. This was the approach that we used in our previous work [9], [1]. Although it offers very good results, the computational burden is extremely high, when dealing with large databases.

In this paper, we will use a template matching method, where the classifier consists of a Nearest Neighbor (NN) classifier using the Mean Square Error (MSE) or the Mean Absolute Difference (MAD) defined as:

$$MSE(\vec{x}, \vec{y}) = \sum_{i=1}^{(N')^2} (x_i - y_i)^2 \quad (6)$$

$$MAD(\vec{x}, \vec{y}) = \sum_{i=1}^{(N')^2} |x_i - y_i| \quad (7)$$

___

Where *N'* is the dimensionality of the vectors that represent faces. According to our previous work [1], we have chosen *N'*=100.

Table 1 shows the obtained results using the ORL database. Each vector, which represents one face, has been obtained with the procedure described in [1].

| Error criterion | Identif. rate | Min(DCF) |
|---|---|---|
| MSE | 91% | 5.84% |
| MAD | 92.5% | 6.28% |

Table 1. Recognition rates for ORL database.

The MAD criterion is faster and simpler to work out. Thus, we have chosen the MAD criterion in our simulations with FERET database.

**Color versus black & white images**

ORL database is in black and white format. Thus, it is not possible to check the relevance of color information using this database. However, FERET database is in RGB format, and the relevance of color can be established. Common sense says that color information is certainly useful for face (pattern recognition). However, common sense is based on our own experience, where we have been "trained" to classify patterns using color information, and our skill is slightly degraded without color information. In fact, Oliver Sacks, a professor of neurology at the Albert Einstein College of Medicine, reports in one of his books [11] some interesting experiences about human recognition and color in special situations:

*"I recently heard of an achromatopic botanic in England said to be even better than color normals at swiftly identifying ferns and other plants in woods, hedgerows, and other almost monochromatic environments. Similarly, in World War II, people with severe red-green colorblindness were pressed into service as bombardiers, because of their ability to "see through" colored camouflage and not be distracted by what would be, to the normal sighted, a confusing and deceiving configuration of colors. One veteran of the Pacific theater reports that colorblind soldiers were indispensable in spotting the movement of camouflaged troops in the jungle. (All of these things may also be clearer to color normals at twilight)."*

| Input signal | Identif. | Min(DCF) |
|---|---|---|
| R | 73.08% | 5.60% |
| G | 68.55% | 5.99% |
| B | 65.63% | 6.09% |
| Y | 69.46% | 5.76% |
| Score fusion: R+G+B | 71.57% | 5.31% |
| Score fusion: 0.3R+0.59G+0.11B | 70.97% | 5.45% |

Table 2. Recognition rates using FERET database.

We have used several data fusion [8] strategies:
a) Feature level data fusion: R, G and B are combined in order to get the luminance of each pixel: Y=0.3R+0.59G+0.11B. Thus, recognition is performed using luminance. This result is comparable to ORL images.
b) Score level data fusion: We work out the recognition rates for each color separately, and we also provide results for score fusion.

Table 2 summarizes the obtained results for FERET database.

## 5. CONCLUSIONS

Our experiments show that:
- Experimental results obtained with ORL are statistically significant, because we get similar performance with a larger database (FERET).
- Color information provides minor improvements when compared with luminance images.
- Proposed method is not suitable for face identification in a large size database (around 1000 users), although it can work reasonably good with small size databases (around 40 people).
- Proposed method is suitable for face verification

## REFERENCES

[1] Faundez-Zanuy, Marcos "Face recognition in a transformed domain". 37º IEEE International Carnahan Conference on Security Technology. Pp.290-297. Taipei (Taiwan), October 2003.

[2] M. Turk M. & A. Pentland, "Eigenfaces for Recognition" Journal Cognitive Neuroscience, Vol. 3, nº 1 pp 71-86, Massachusetts Institute of Thecnology.1991.

[3] "Color FERET. Facial Image Database.", Image Group, Information Access Division, ITL, National Institute of Standards and Technology. Oct. 2003.

[4] Bolle, R., Connell, J., Pankanti, S., Ratha, N., Senior, A.: Guide to biometrics. Springer professional computing, New York 2004

[5] Guyon, I., Makhoul, J., Schwartz, R., and Vapnik, V.: 'What size test set gives good error rate estimates?', IEEE Trans. Pattern Anal. Mach. Intell., 1998, 20, (1), pp. 52–64.

[6] Samaria F., Harter A. "Parameterization of a stochastic model for human face identification". 2nd IEEE Workshop on Applications of Computer Vision December 1994, Sarasota (Florida).

[7] Zhou S.K. "Face recognition using more than one still image: what is more?. Lecture Notes In Computer Science LNCS 3338, pp.212-223, A. Z. Li et al. Ed., Sinobiometrics. Springer Verlag 2004

[8] Faundez-Zanuy Marcos., "Data fusion in biometrics" IEEE Aerospace and Electronic Systems Magazine. Vol.20 nº 1, pp.34-38, January 2005.

[9] Espinosa V., Faundez-Zanuy Marcos "Face identification by means of a neural net classifier" 33º IEEE International Carnahan Conference on Security Technology. Pp.182-186. Madrid, 1999

[10] Martin A., Doddington G., Kamm T., Ordowski M., and Przybocki M., "The DET curve in assessment of

___

detection performance", V. 4, pp.1895-1898, European speech Processing Conference Eurospeech 1997

[11] Sacks O., "An anthropologist on Mars". First Vintage books edition, USA, Feb. 1996